\documentclass[1p]{elsarticle}
\usepackage{lineno,hyperref}
\usepackage{amsmath, array}
\usepackage{subfig}
\usepackage{amssymb}
\usepackage{multirow}
\usepackage{makecell}
\usepackage[ruled,vlined]{algorithm2e}
\usepackage[utf8]{inputenc}
\SetKwComment{Comment}{$\triangleright$\ }{}

\journal{Journal of Pattern Recognition}
\begin{document}

\begin{frontmatter}

\title{Learning Multi-level Weight-centric Features for Few-shot Learning}

\author{Mingjiang Liang}
\author[1]{Shaoli Huang}
\cortext[mycorrespondingauthor]{Corresponding author}
\ead{shaoli.huang@sydney.edu.au}
\address[1]{University of Sydney}
\author[2]{Shirui Pan}
\address[2]{Monash University}
\author[3]{Mingming Gong}
\address[3]{University of Melbourne}
\author[4]{Wei Liu}
\address[4]{University of Technology Sydney}



\begin{abstract}
Few-shot learning is currently enjoying a considerable resurgence of interest, aided by the recent advance of deep learning. Contemporary approaches based on weight-generation scheme delivers a straightforward and flexible solution to the problem. However, they did not fully consider both the representation power for unseen categories and weight generation capacity in feature learning,  making it a significant performance bottleneck. This paper proposes a multi-level weight-centric feature learning to give full play to feature extractor's dual roles in few-shot learning. Our proposed method consists of two essential techniques: a weight-centric training strategy to improve the features' prototype-ability and a multi-level feature incorporating a mid- and relation-level information.  The former increases the feasibility of constructing a discriminative decision boundary based on a few samples. Simultaneously, the latter helps improve the transferability for characterizing novel classes and preserve classification capability for base classes. We extensively evaluate our approach to low-shot classification benchmarks.  Experiments demonstrate our proposed method significantly outperforms its counterparts in both standard and generalized settings and using different network backbones.
\end{abstract}


\end{frontmatter}


\section{Introduction}

Despite remarkable success made in visual recognition tasks \cite{zhou2014learning,he2016deep}, deep learning models generally lack versatility and extendability, hindering their applicability in practice. For instance, being data-hungry to learn massive parameters, deep neural networks often fail to work well in data-scarce environments. Besides, a trained model's prediction domain is usually not expandable unless re-executing the training process. In response to these deficiencies, there has been increasing efforts devoted to few-shot learning (FSL) \cite{finn2017model,hariharan2017low,qi2018low,yang2018learning,snell2017prototypical,wang2018low,oreshkin2018tadam,lifchitz2019dense,lee2019meta}.   FSL refers to a technique that exploits knowledge from base-class data (provided auxiliary training set) to allow models to understand new concepts from only a few examples.  Existing approaches to this problem mainly consist of meta-learning and weight-generation based frameworks. The former focuses on learning a meta-learner from base-class data to facilitate learning a new-task learner. Although meta-learning approaches achieve great success, they often require sophisticated training procedures and are difficult to extend to generalized few-shot learning (GFSL) settings. 

\begin{figure}
\begin{center}
  \includegraphics[width=0.7\linewidth]{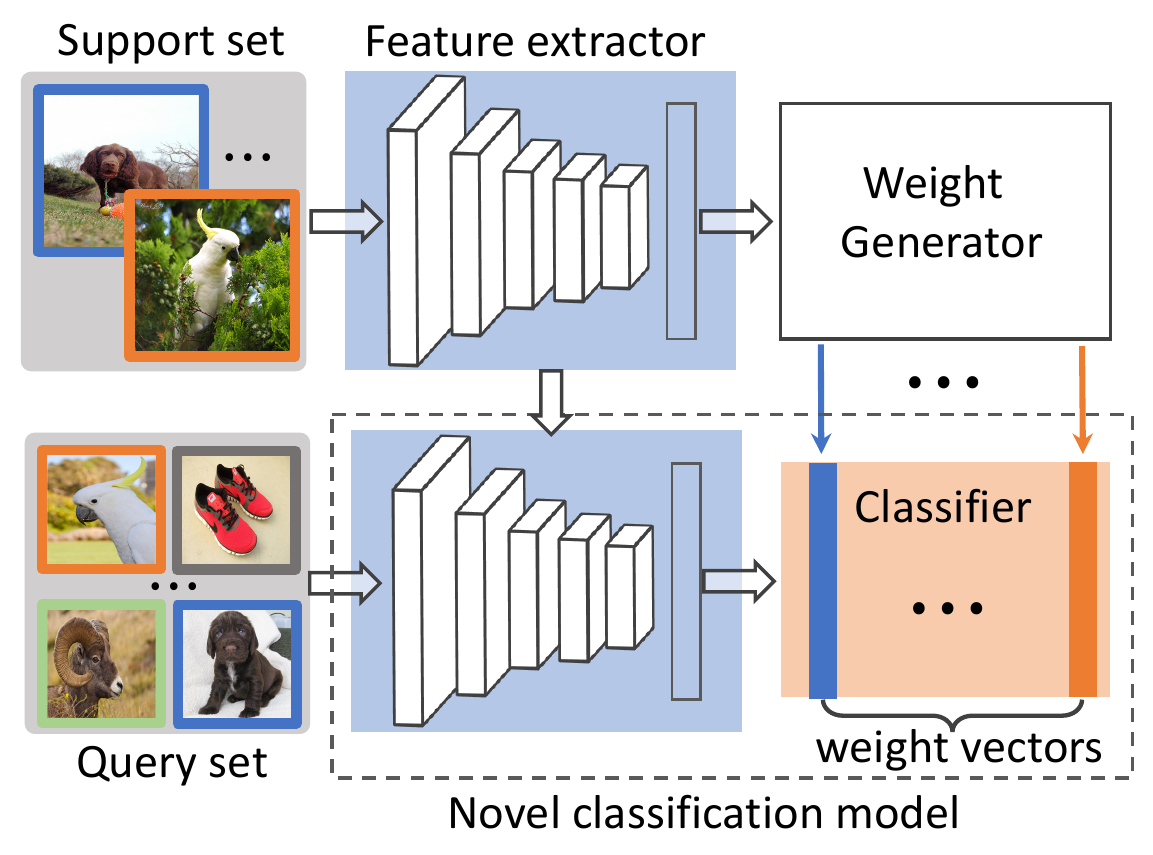}
\end{center}
   \caption{A general framework of weight generation methods for few-shot learning.  Learning a good feature extractor plays a vital role in this framework,  as it is used for novel models to extract image features and generate classifier weights for new categories.}
\label{fig:title}
\end{figure}

By contrast, the weight-generation framework delivers a more straightforward and flexible solution. This type of method first learns an embedding space from base-class data and then utilizes the embedding of support set  (novel-class training samples ) to construct the corresponding classifier weights(as illustrated in Fig.~\ref{fig:title}.  This learning regime simplifies the few-shot learning problem by mainly focusing on feature learning and weight generation, enabling a trained model's extendability. 

 In the framework, feature learning is crucial due to its dual-use mechanism (representing images and constructing classifiers). However, existing methods learn a feature extractor without considering three essential issues associated with its dual-functionality: representation transferability, base-class memorability, and prototype-ability.  The transferability refers to whether the learned representation from base-class data is transferable to the novel-class data. Recent works \cite{vinyals2016matching,qi2018low,gidaris2018dynamic,wang2018low} mainly extract features from the last Conv layer of deep models, leading to less transferability of the feature extractor. This can be attributed to the fact that higher layer activations with higher specialization to base-class data are less transferable to novel-class \cite{yosinski2014transferable}. The memorability and prototype-ability are more related to the quality of the generated classifier weights. A GFSL model requires preserving the classification performance for the base-class data. This requirement necessitates base-class memorability to prevent novel-class weights from classifying base-class data to novel classes. The prototype-ability refers to the feature extractor's capacity in allowing few-shot examples to approximate their class-specific prototype. Current methods attempt to complement these two capabilities by learning a weight generation network.
Nevertheless,  similar to meta-learning approaches, they need training a new task-specific learner for weight generation, limiting the flexibility to construct few-shot classification models. Besides, the weight generator learns the required information from extracted features but fails to access more information through the feature learning stage. To sum up, the existing weight-generation methods do not fully consider the feature extractor's dual-capacity in FSL, which may be a bottleneck of performance.

In this paper, we propose a multi-level weight-centric feature extractor to complement the capacity of current weight-generation methods.  We first introduce a weight-centric training strategy to increase the possibility that each sample can approximate its category prototype. Specifically, we fix the classifier weights in the latter learning stage and then enforce samples closer to their corresponding classifier weight in the embedding space.  Besides, we build the multi-level feature by incorporating a mid-level and relation-level learning branch with high-level feature learning. The mid-level learning branch extracts mid-level features from intermediate layers while the relation-level one obtains category-relation information from softening predictions.  We finally integrate the multi-level information extraction and the weigh-centric strategy into an overall feature learning framework. 

Our proposed method ensures the feature extractor's comprehensiveness for advancing few-shot learning. On the one hand, the weight-centric strategy reduces the intra-class variance, improving feature representation generalization. It also pushes data points that are closer to the hyperplane far way. This effect indirectly achieves larger margin classification-boundaries,  increasing the feasibility of constructing a discriminative decision boundary based on a few samples. On the other hand,  the mid-level features are more transferable \cite{taigman2015web} to novel classes, and the relation-level representation exhibits higher-level abstraction and more specific to base categories. Therefore,  by jointly representing images using these two additional information sources,  the resulting model has higher transferability for characterizing novel classes and better preserves classification capability for base classes. 

We extensively evaluate our approach on two low-shot classification benchmarks in both standard and generalized FSL learning settings. Experiments show that our proposed method significantly outperforms its counterparts in both learning settings and using different network backbones. We also demonstrate that the mid-level features exhibit strong transferability even in a cross-task environment and the relation-level features help preserve base-class accuracy in the generalized FSL setting.

The contribution of this paper can be summarized as :
\begin{itemize}
    \item We propose a weight-centric learning strategy that helps reduce the inter-class variance of novel-class data.
    \item we propose a multi-level feature learning framework, which demonstrates its strong prototype-ability and transferability even in a cross-task environment for few-shot learning.
    \item We extensively evaluate our approach on two low-shot classification benchmarks in both standard and generalized FSL learning settings. Our results show that the mid-level features exhibit strong transferability even in a cross-task environment while the relation-level features help preserve base-class accuracy in the generalized FSL setting
\end{itemize}

\section{Related work}

Recently proposed approaches to few-shot learning problem can be roughly divided into meta-learning based \cite{ravi2017optimization,finn2017model,hariharan2017low,yoon2019tapnet,xing2019adaptive,bertinetto2018meta} and weight-generation based approaches \cite{koch2015siamese,vinyals2016matching,qi2018low,gidaris2018dynamic,li2019few}.

\textbf{Meta-learning based methods} tackle the few-shot learning problem by training a meta-learner to help a learner can effectively learn a new task on very few training data \cite{liu2018transductive,qiao2018few,liu2018transductive,rusu2018meta,finn2017model,hariharan2017low,sun2019meta}. Most of these methods are normally designed based on some standard practices for training deep models on limited data, such as finding good weights initialization \cite{finn2017model} or performing data augmentation \cite{hariharan2017low} to prevent overfitting. For instance, Finnn et al. \cite{finn2017model} propose to learn a set of parameters to initialize the learner model so that it can be quickly adapted to a new task with only a few gradient descent steps; \cite{hariharan2017low} deal with the data deficiency in a more straightforward way, in which a generator is trained on meta-training data and used to augment feature of novel examples for training the learner. Another line of work addresses the problem in a "learning-to-optimize" way \cite{rusu2018meta,ravi2017optimization}. For example, Ravi et al. \cite{ravi2017optimization} train an LSTM-based meta-learner as an optimizer to update the learner and store the previous update records into the external memory. Though this group of methods achieves promising results, they either require to design complex inference mechanisms \cite{fei2006one} or to further train a classifier for novel concepts \cite{ravi2017optimization,finn2017model}. Our work focuses on learning a feature extractor with dual functions (ie feature representation and classifier weight generation) for FSL problems. Therefore, the major difference from meta-learning techniques is that our method only needs to learn a base model and can construct new models directly using sample features.

\textbf{Weight-genration based approaches} mainly learn an embedding space, in which images are easy to classify using a distance-based classifier such as cosine similarity or nearest neighbor. To do so, Koch et al. \cite{koch2015siamese} trains a Siamese network that learns a metric space to perform comparisons between images. Vinyals et al. \cite{vinyals2016matching} propose Matching Networks to learn a contextual embedding,with which the label of a test example can be predicted by looking for its nearest neighbors from the support set. Prototypical networks \cite{snell2017prototypical} determine the class label of a test example by measuring the distance from all the class means of the support set. Since the distance functions of these two works are predefined, \cite{yang2018learning} further introduce a learnable distance metric for comparing query and support samples. 

The most related methods to ours are \cite{qi2018low,gidaris2018dynamic,chen2018closer}. These approaches learn a feature representation by a cosine softmax loss, allowing a few novel examples to construct the classifier.  
Our proposed method differs from them in two folds. First, they only learn a single level of representation, resulting in a limited representation capability,  while ours constructs a multi-level model that considers multiple knowledge sources. Furthermore, those methods do not explicitly consider the prototype-ability ( the ability to approximate the corresponding prototype by one or several sample features) in learning the feature extractor. In contrast, we introduce a weight-centric learning strategy that makes it more feasible to construct classifier weights from a few samples.


\subsection{Analyzing the transferability of ConvNets.} Deep learning models are quite data-hungry but nonetheless transfer learning have been proven highly effective to avoid over-fitting when training larger models on smaller datasets\cite{donahue2014decaf,zeiler2014visualizing,sermanet2013overfeat}. These findings raise interest in studying the transferability of deep models features in recent years. Yosinski et al. \cite{yosinski2014transferable} experimentally show how transferable of each layer by quantifying the generality versus specificity of its features from a deep ConvNet, and suggest that higher layer activations with higher specialization to source tasks are less transferable to target tasks. Pulkit et al. \cite{agrawal2014analyzing} investigates several aspects that impact the performance of ConvNet models for object recognition. Hossein et al. \cite{azizpour2015generic} identifies several factors that affect the transferability of ConvNet features and demonstrates optimizing these factors aid transferring task. However, these works mainly explore the transferability and generalization ability of ConvNet features in terms of target datasets where the training samples are much more than the few-shot setting. In this work, we investigate the capacities of the intermediate layer, last feature layer, and softmax logits to perform few-shot learning tasks.

\begin{figure}[t]
\begin{center}
  \includegraphics[width=0.85\linewidth]{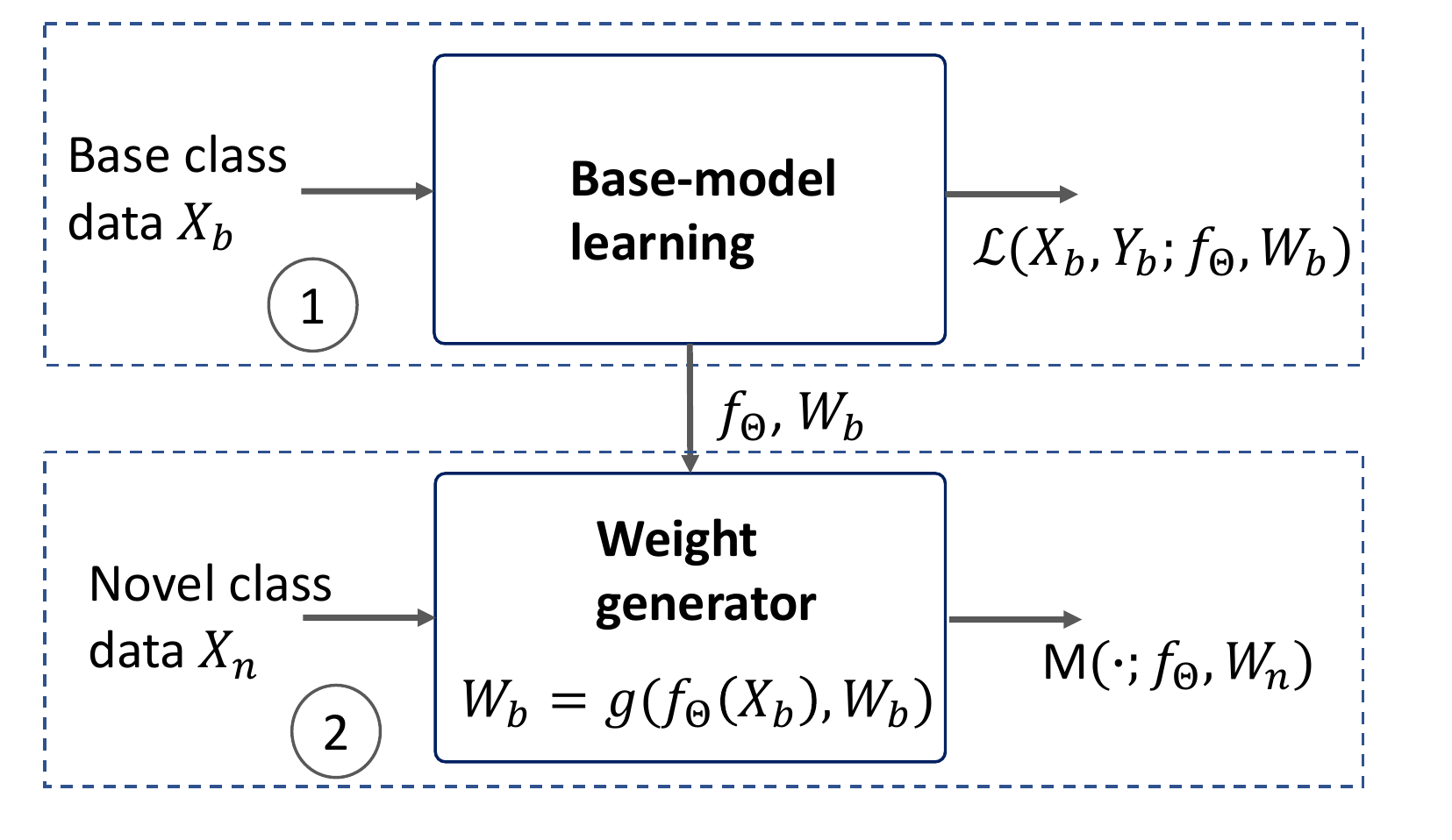}
\end{center}
   \caption{ A general weight-generation-based framework for few-shot learning. Here, $\mathcal{L}$ is the loss function for learning a base model on base-class data. $f_{\Theta}$ and $W_b$ are the feature extractor and classifier weights of the base model. $g(\cdot)$ is weight generator which can be defined or learned from data. $M(\cdot)$ is novel model built for novel categories. }
\label{fig:general}
\end{figure}

\begin{figure*}[t]
\begin{center}
  \includegraphics[width=0.95\linewidth]{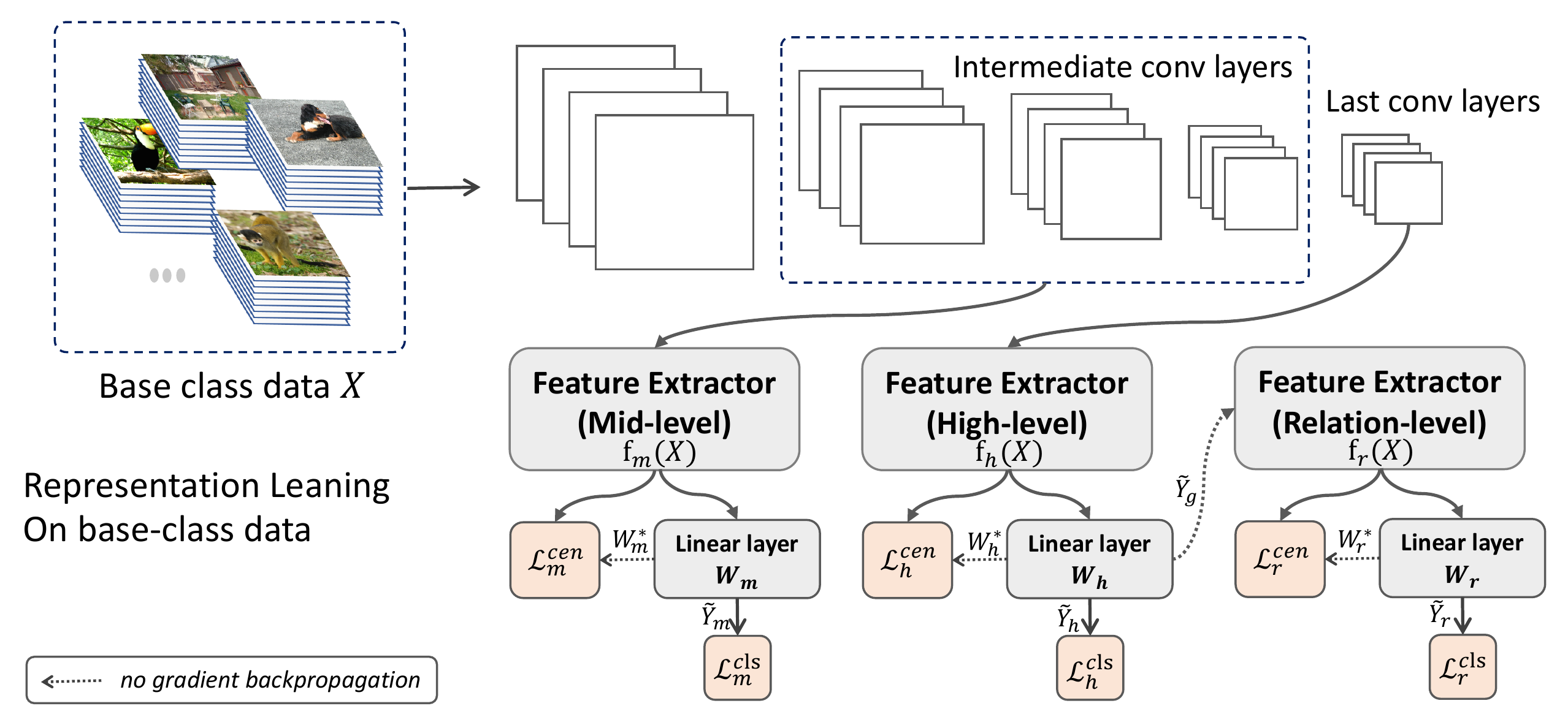}
\end{center}
   \caption{An overview of our learning framework for representation learning. The framework first construct three levels of feature representations namely mid-level,high-level,and relation-level. For forwarding the networks, the outputs of intermediate layer outputs are detached and fed to the mid-level feature extractor, the output of the last conv layer is forwarded to high-level feature extractor, and the prediction logits of high-level branch are detached and input and sent to the relation-level feature extractor. The three feature extractors are first trained to converge with the classification loss, and then are further fine-tuned with both the classification and the weight-centric loss.}
\label{fig:overall}
\end{figure*}


\section{Methodology}

In this section, we first introduce some general notation used throughout the paper. We first briefly review a general weight-generation-based framework for few-shot learning. We further introduce our method for learning base models. Finally, we describe how to utilize these base models in few-shot learning.


\subsection{Notation}

Let $f_{\Theta}(\cdot) \in \mathbb{R}^d$ be a feature extractor parameterized by $\Theta$ and  $W \in \mathbb{R}^{d\times c}$  be a weight matrix of a linear classifier. Here, $d$ is the dimension of the output feature and $c$ is the number of labels for the classification task. We further define $M(\cdot)$ as a neural network classification model, such that $M(f_{\Theta}(x),W) =W^Tf_{\Theta}(x)$ given an input image $x$.  We denote the training set $D_{train}$ and the test set $D_{test}$. Slightly different from the general classification setting, few-shot learning train a model $M(\cdot)$ on the training data that consists of a base- and novel-class dataset, that is $D_{train} = D_{train}^b \cup D_{train}^n$. Here,  $D_{train}^b=\{(x_i,y_i),y_i\in Y^b\}_{i=1}^{N^b}$ is an abundant dataset  while $D_{train}^n=\{(x_i,y_i),y_i\in Y^n\}_{i=1}^{N^n}$ contains very few samples for each label; $Y^b$ and $Y^n$ refers to two different label spaces and  $Y^b \cap Y^n = \emptyset$.  We further denote the weight matrices $W^b$ and $W^n$ which are corresponding to $Y^b$ and $Y^n$ respectively.

\subsection{Weight-generation-based framework}
\label{sec:generalframework}
 Weight-generation-based approaches have gained increasing attention in recent years, due to its simplicity and flexibility. The general framework for these methods usually consists of two stages: base-model learning and weight generation. As shown in Fig.\ref{fig:general}, this framework first learns a classification base-model on base-class dataset. In the second stage, based on the feature extractor $f_{\Theta}(\cdot)$ and classifier weights $W^b$ of the base-model, a weight generator $g_{\phi}(.)$ is used to infer the weight vector $w$ given training set $X^y=\{x^y_1,...,x^y_k\}$. Here, the label $y$ is in an unseen label space $Y^n$ and $k$ is usually a small number. In recent literature, there are two typical weight generators : average-based $w_{avg}=g^{avg}(f_{\Theta}(X^y))$ and attention-based $w_{att}=g^{att}_{\phi}(f_{\Theta}(X^y),W^b)$. The former simply compute the mean of the normalized features of training samples, which is expressed as:

\begin{equation}
    w_{avg} = \frac{1}{k}\sum_{i=1}^kz_i,
\end{equation}
where $z_i$ is a $L_2$ norm of the feature vector $f_{\Theta}(x^y_i)$.

The second one employs an attention-based mechanism to exploit both the sample features and the base-class weights in generating the novel-class weights. The weight computation for an unseen label is expressed as:

\begin{equation}
    w_{att} = \phi_{avg}\odot w_{avg}+\phi_{att}\odot (\frac{1}{k}\sum_{i=1}^k\sum_{b=1}^{K_b}Att(\phi_qz_i,k_b)\cdot w_b
\end{equation}
where $odot$ is the Hadamard product,$\phi_{avg},\phi_{att},\phi_q$ are learnable parameters,$Att(., .)$ is an attention kernel, and $\{k_b \in R^d\}_{b=1}^{K_b}$ is a set of $K_b$ learnable keys.


\subsection{Multi-level Weight-centric (MLWC) Representation Learning }
\label{sec:representation}
Fig.~\ref{fig:overall}~provides an overview of our proposed method.
The method mainly consists of two techniques: a multi-level feature extractor and a weight-centric feature learning strategy. The former aims to explicitly enforce each single sample feature vector closer to its corresponding classifier weight. Specifically, we construct three levels of feature representations namely mid-level, high-level, and relation-level. The mid-level representation captures more subtle discriminative patterns, such as subordinary components of object parts, while the high-level encodes more holistic information. The relation-level is designed to describe the input's category structural relations, like how the input image relates to other categories.
 The second technique intends to obtain multiple representations that encode different levels of semantic information. Overall, the multi-level extractor improves the representation ability by considering multiple sources of information, and the weight-centric strategy increases the feasibility of generating classifier weights from few-shot sample features. These two techniques can seamlessly join together to provide a simple and effective solution to few-shot learning problems.

\begin{figure}[t]
\begin{center}
      \includegraphics[width=0.7 \linewidth]{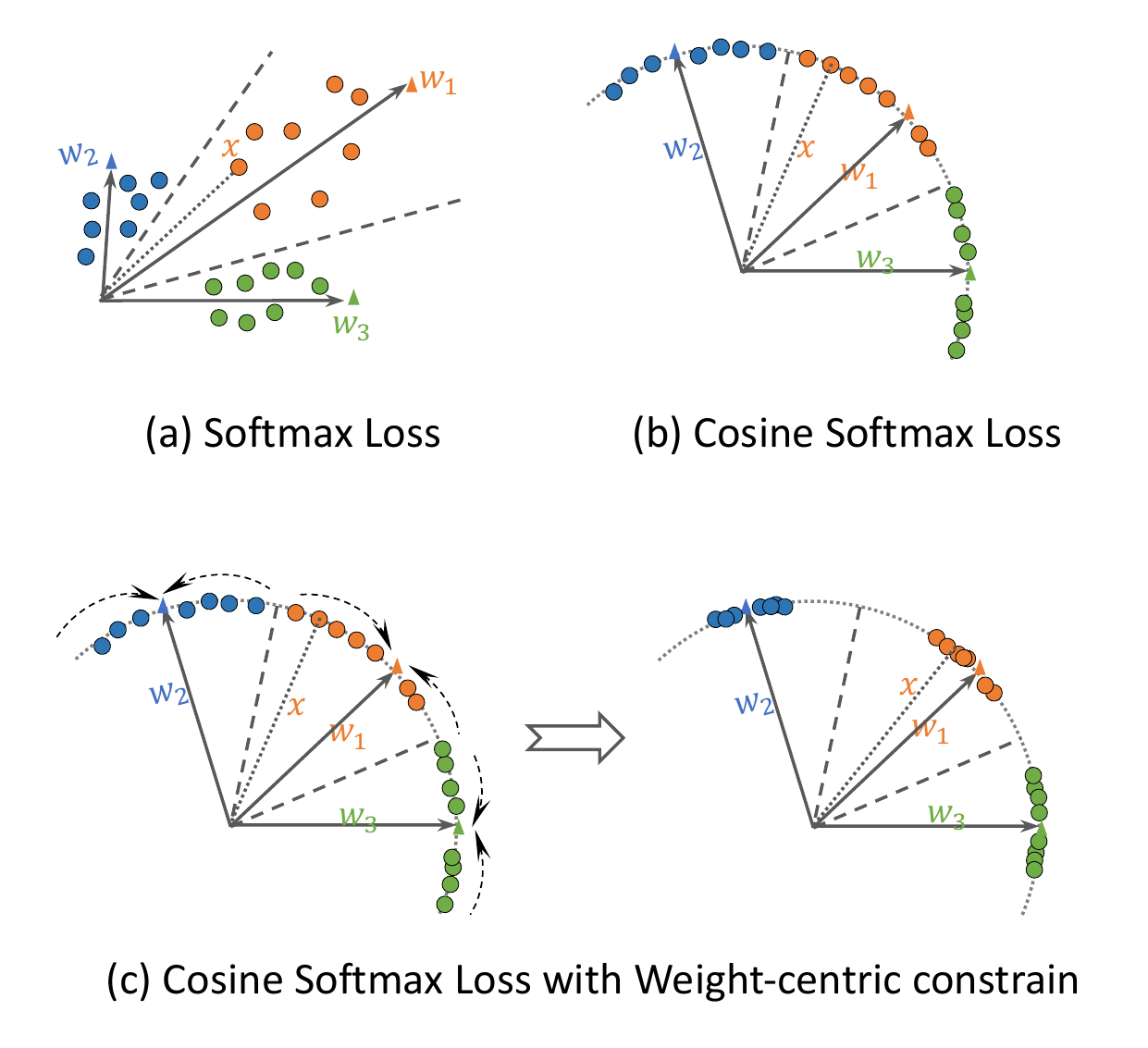}
\end{center}
   \caption{A geometry interpretation for learning feature space with different loss functions. }
\label{fig:centricloss}
\end{figure}

\subsubsection{Learning weight-centric Feature Embedding}

In this subsection, we first review cosine softmax loss for few-shot learning.  We then introduce our proposed weight-centric embedding learning strategy. This strategy can be incorporated with cosine softmax loss to facilitate the subsequent step of generating weights from few-shot training examples. 

\noindent\textbf{Cosine Softmax Loss}.
In standard classification framework, Softmax Loss is usually adopted for supervised learning. It generally refers to a Softmax Activation plus a Cross-Entropy Loss.  Given an input $(x_i,y_i)$, the softmax loss function is expressed as:
\begin{equation}
\ell_s(x_i,y_i) = -log(\frac{exp(w_{y_i}^Tf_{\Theta}(x_i))}{\sum_j{exp(w_j^Tf_{\Theta}(x_j))}}),
\end{equation}
where $f_{\Theta}(\cdot)$ is the feature extractor and $w_j$ is the $j^{th}$ column of the the weight matrix $W$ of the classifier layer.

However, recent works show the softmax loss fails to learn a feature extractor that generalizes well to unseen categories \cite{qi2018low,gidaris2018dynamic}.
As discussed previously, the feature extractor of the base model is used to generate weights of novel categories.  the weak transferability of feature extractor will consequently degrade the performance of the novel model.  To ease this issue, \cite{qi2018low,gidaris2018dynamic} propose to adopt cosine softmax loss in learning the base model. Compared with softmax loss, Cosine softmax loss applies  $l_2$-normalization on both the feature vector and the weight vector before the loss calculation, which is expressed as:
\begin{equation}
	\tilde{w}_j = \frac{w_j}{\|w_j\|}, \tilde{f}_{\Theta}(x_i) = \frac{f_{\Theta}(x_i)}{\|f_{\Theta}(x_i)\|}.
\end{equation}

\begin{algorithm*}
\SetAlgoLined
\SetKwInOut{Input}{Input}
\SetKwInOut{Output}{Output}
\Input{Base-class Training data $\{X,Y\}$,feature extractor with parameters of $\Theta$},linear classifier weights $\mathcal{W}$.

\Output{Updated $\Theta$ and $\mathcal{W}$}

 Initialize parameters  $\Theta$ and $\mathcal{W}$\
 

  \While{$\mathcal{L}_{cs}$ not converge}{  
  Sample a minibatch of $m$ examples from the training set $\{x^{(1)},...,x^{(m)}\}$ with corresponding targets $y^{(i)}$\;
   Compute gradient: $g_{\Theta} \longleftarrow \frac{1}{m} \bigtriangledown_{\Theta}\sum_{i}\mathcal{L}_{cls}(x^{(i)},y^{(i)};\Theta,\mathcal{W})$ \Comment*[r]{$\mathcal{L}_{cs}$ is computed using eq.\ref{eq:cosineloss}} 
    Compute gradient: $g_{\mathcal{W}} \longleftarrow \frac{1}{m} \bigtriangledown_{\mathcal{W}}\sum_{i}\mathcal{L}_{cls}(x^{(i)},y^{(i)};\Theta,\mathcal{W})$ \;
    update $\Theta$ and $\mathcal{W}$ \;
   }
   $\mathcal{W}^{*} \longleftarrow \mathcal{W}$ \Comment*[r]{Frozen classifier weights} 
   \While{$\mathcal{L}_{centric}$ and $\mathcal{L}_{cls}$ not converge}{
   Sample a minibatch of $m$ examples from the training set $\{x^{(1)},...,x^{(m)}\}$ with corresponding targets $y^{(i)}$ \;
 
   Compute gradient: $g_{\Theta} \longleftarrow \frac{1}{m} \bigtriangledown_{\Theta}\sum_{i}(\mathcal{L}_{cls}(x^{(i)},y^{(i)};\Theta,\mathcal{W}^{*})+\mathcal{L}_{cen}(x^{(i)};\Theta,\mathcal{W}^{*}))$\;
   
   update $\Theta$\;
   }
 \caption{Learning weight-centric features}
 \label{alg:alg1}
\end{algorithm*}

\begin{figure*}[t]
\begin{center}
  \includegraphics[width=1\linewidth]{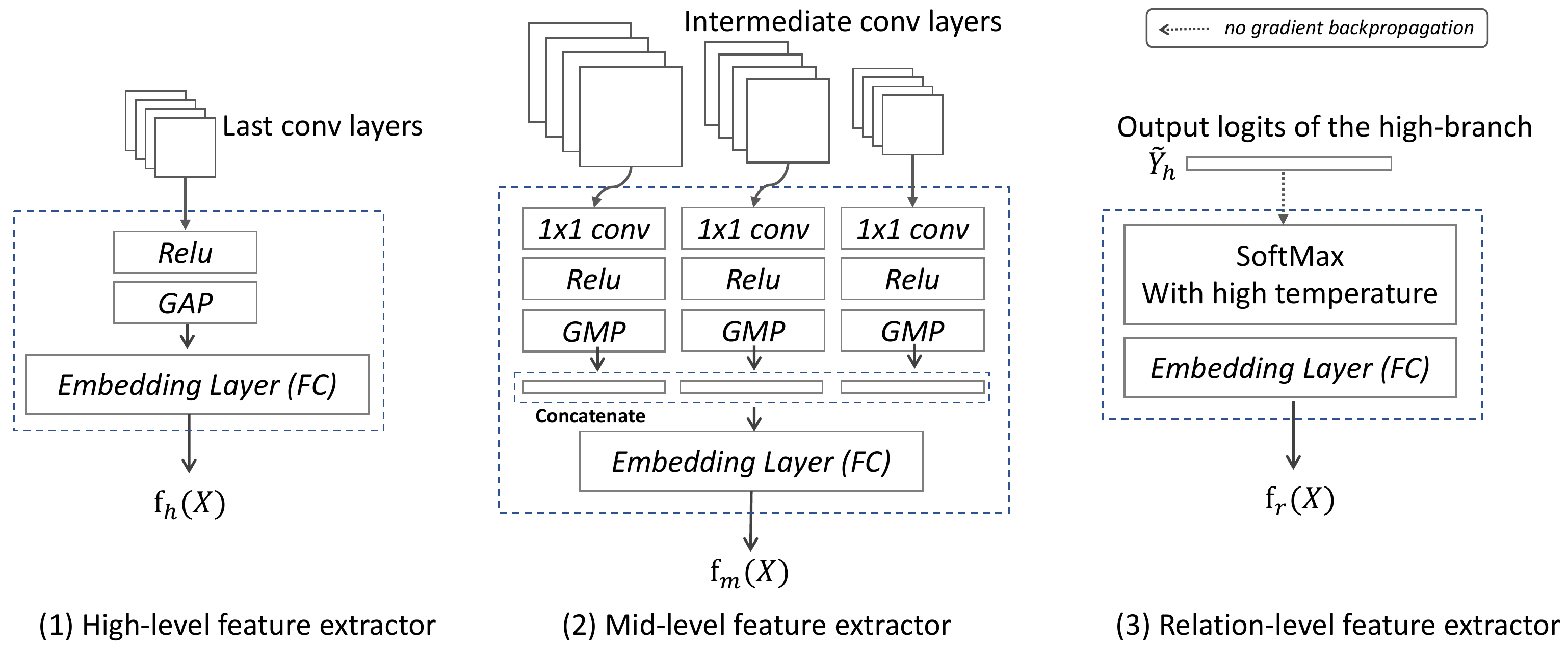}
\end{center}
   \caption{Sub-network structures for different levels of feature extractor. }
\label{fig:fea}
\end{figure*}

This normalization step will cause the softmax function to fail to produce a one-hot categorical distribution, making the neural networks hard to converge. As suggested in \cite{qi2018low}, a simple solution to this is to introduce a trainable scale factor $s$ into to the softmax function. Thus, the cosine softmax loss function is expressed as:
\begin{equation}
\ell_{cs}(x_i,y_i;\Theta,W) = -log(\frac{s\cdot exp(\tilde{w}_{y_i}^T\tilde{f}_{\Theta}(x_i))}{\sum_j{exp(\tilde{w}_j^T\tilde{f}_{\Theta}(x_i))}}).
\end{equation}
Based on this loss function, \cite{qi2018low,gidaris2018dynamic} learn the feature extractor by minimizing the cost function
\begin{equation}
\mathcal{L}_{cs}= \frac{1}{N}\sum_i^N(\ell_{cs}(x_i,y_i;\Theta,W)) + \lambda R(W),
\label{eq:cosineloss}
\end{equation}
where $\lambda R(W)$ is a weight $L_2$ regularization term.

\begin{figure*}[t]
\begin{center}
  \includegraphics[width=1\linewidth]{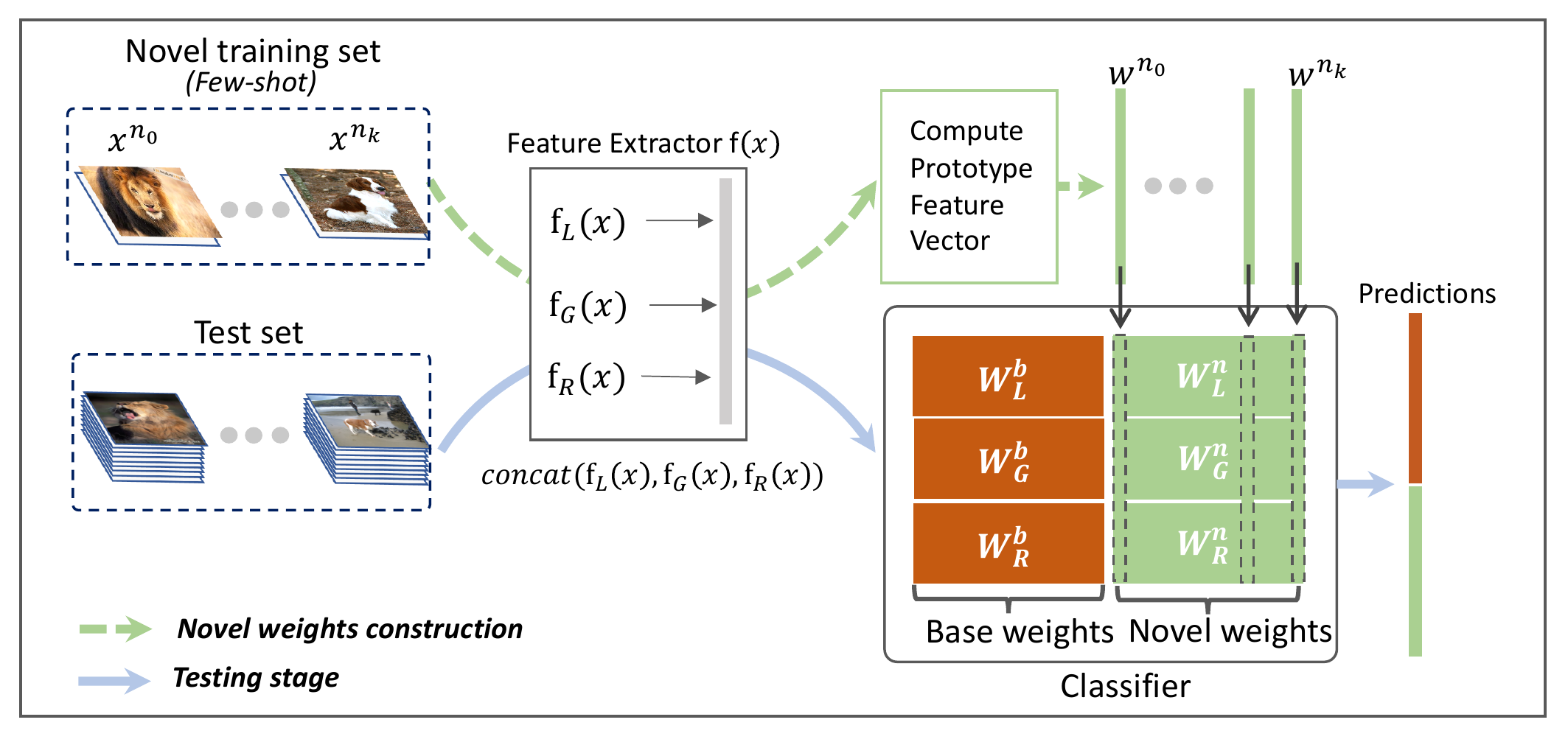}
\end{center}
   \caption{Utilizing the base models to construct classification models in few-shot learning. We first combine the base modes to obtain a multi-level feature extractor and a base-class weight matrix. Then the feature extractor is used to produce a novel-class weight matrix. Finally, we can construct classification models for few-shot learning by integrating the feature extractor, base- and novel-class weight matrix. }
\label{fig:fewshot}
\end{figure*}

\noindent\textbf{Weight-Centric feature learning}. As illustrated in Fig.\ref{fig:centricloss} (a) and (b), learning with cosine softmax loss reduces intra-class variations by comparison with original softmax loss. Thus it increases the feasibility to characterize an unseen concept with few-shot examples.  On the other hand, it also decreases variations between inter-class samples, which will diminish the novel model's discriminating ability when using weights generated from one or few-shot examples. To address this issue, we propose a weight-centric constraint to improve the feature learning using the cosine softmax loss. The proposed constraint forces each feature point close to its corresponding class weight point, which can be formularized as 

\begin{equation}
	\ell_{cen}(x_i,w^*_{y_i};\Theta) = \parallel f_{\Theta}(x_i)- w^{*}_{y_i}\parallel^2,
	\label{eq:centric}
\end{equation}

where $w^*_{y_i}$ represents the sample $x_i$'s corresponding class weight vector, which specifically refers to the ${{y_i}^{th}}$ column of a constant matrix $W^*$. Noted we obtain $W^*$  from the classifier layer after first training the model to converge using the cost function $\mathcal{L}_{cs}$.
To couple the constraint with cosine softmax loss, we also apply $l_2$-normalization on both the feature vector and the weight vector. Thus, the weight-centric constraint can be rewritten as 
\begin{equation}
	\ell_{cen}(x_i,w^*_{y_i};\Theta) = \parallel \frac{f_{\Theta}(x_i)}{\|f_{\Theta}(x_i)\|}- \frac{w^{*}_{y_i}}{\|w^{*}_{y_i}\|}\parallel^2.
	\label{eq:centric}
\end{equation}

By integrating the cosine softmax loss and the weight-centric constraint, we now have the cost function $\mathcal{L}$.
\begin{equation}
	\mathcal{L} = \begin{cases}\mathcal{L}_{cs} &\mathcal{L}_{cs} > \epsilon \cr \mathcal{L}_{cen}+\mathcal{L}_{cs} ,  &otherwise,\end{cases}
	\label{eq:finalloss}
\end{equation}
where $\mathcal{L}_{cen} = \frac{1}{N}\sum_i^N(\ell_{cen}(x_i,w^*_{y_i}))$ and $\mathcal{L}_{cs} > \epsilon$ means that the stopping criteria is not met when training with loss $\mathcal{L}_{cs}$. Since the $\mathcal{L}_{cen}$ required $W^*$ as input, we optimize the cost function using a two-stage algorithm which is detailed in Algorithm.\textbf{1}.

As illustrated in Fig.\ref{fig:centricloss} (c), the weight-centric constraint push samples closer to their corresponding classifier weights, which brings two advantages. First, it enforces the neural network to learn a feature space with smaller intra-class variance. Moreover, the constraint also implicitly drives samples far away from the decision boundary. This increases the feasibility of constructing a discriminative decision boundary based on a small number of samples. 

\subsubsection{Multi-level Feature Extractor}

A good representation of generalized few-shot learning is it can generalize well to novel concepts while maximizing its original ability to discriminate base categories. A single high-level of feature representation usually has limited capacity to meet these criteria simultaneously. In this subsection, we introduce two additional levels of representation namely mid- and relation-level to complement the representative capacity of high-level representation. 

\noindent\textbf{High-level feature extractor} is a common practice in most existing few-shot learning methods. As illustrated in Fig.~\ref{fig:fea}~(1),  It takes inputs from the last convolutional layer and then maps them into an embedding space after applying global-average pooling. This design results in the extracted features naturally capture the global visual discriminative patterns, because of the high-level feature abstraction source and the property of the pooling operation.

\noindent\textbf{Mid-level feature extractor} aims to obtain features that focus more on encoding mid-level discriminative patterns. Compared with the high-level features,  it exhibits a  better generalization ability in representing novel concepts but weaker discriminatory power for the base concepts.   This can be attributed to the fact that it tends to abstract information that is less specific to the base concepts.  A naive scheme to learn mid-level features is to plug an additional global-extractor head on top of the intermediate layers.  However, this solution might still learn features more similar to the high-level ones because of the global average pooling operation, though the input source is switched to the lower layers.  To avoid such undesirable effects, we design the mid-level feature extractor, shown in Fig.~\ref{fig:fea}~(2). Specifically, we insert a 1x1 Conv layer on top of each intermediate layer and employ global-max pooling to prevent the 1x1 Conv layer from learning global abstraction. Lastly, we concatenate all the intermediate-layer features into one and map it into embedding space to form a compact mid-level representation.

\noindent\textbf{Relation-level feature extractor.} As discussed previously, the model's generalization to novel concepts can be improved by incorporating the mid-level representation. However, its ability to classify base classes is degrading when the label space is expanding with more novel classes (some base-class examples might be misclassified to novel classes). Thus, we propose to preserve such ability by encoding more specific information of base classes. Specifically, we introduce another relation-level representation that describes an input using its structural relationships within the base classes. This representation is more specific to base classes than both the high- and mid-level representation. Though it has a poor generalization to novel concepts, it helps strengthen the classification capacity for base classes.  As shown in Fig.~\ref{fig:fea}~(3),  the relation-level extractor takes inputs the predicted logits $\tilde{Y}_{h}$ from the high-level branch. Here, when the $\tilde{Y}_{h}$ from a trained model is fed to a softmax layer, the outputs will tend to be a one-hot vector, which fails to describe the data's structural relation over classes. Therefore, we feed $\tilde{Y}_{h}$  to a softmax function with a high temperature, so that it can encode a richer class structural information of the data. Finally, we use this soften prediction outputs to learn the embedding space that characterizes the similarity of samples according to their categorical distribution.

\noindent\textbf{Jointly Learning multiple feature extractors}. As shown in Fig.~\ref{fig:overall}, we learn the three feature extractors using three classification branches that are all based on a single network backbone. We also apply the weight-centric learning strategy for each branch. Thus, the overall classification loss and weight-centric loss 
\begin{equation}
\begin{aligned}
\mathcal{L}_{cs} &= \mathcal{L}^m_{cs} + \mathcal{L}^h_{cs} + \mathcal{L}^r_{cs}, \\
\mathcal{L}_{cen} &= \mathcal{L}^m_{cen} + \mathcal{L}^h_{cen} + \mathcal{L}^r_{cen}
\end{aligned}
\end{equation} 
respectively. Finally, our overall cost function is obtained by substituting these two equations into Eq.~\ref{eq:finalloss}.

\subsection{Few-shot leaning}
\label{sec:weights}

In the previous section, we describe how our proposed method learns base models on the base-class dataset. In this section, we describe how to utilize these base models to perform few-shot learning. This procedure mainly consists of two operations: model combination and weight generation, which are detailed in the following.

\noindent\textbf{Model combination.} After training the base models using our proposed method, we have three base models $M(f_m(x),W^b_m)$,$M(f_h(x),W^b_h)$,and $M(f_r(x),W^b_r)$, which denote the mid-, high-, and relation-level classification model respectively. We simply combine them into a single model $M(f_C(x),W_C^b)$ by concatenating their normalized features and classifier weights separately. Here, $f_C(x)=concat(\frac{f_m(x)}{\|f_m(x)\|},\frac{f_h(x)}{\|f_h(x)\|},\frac{f_r(x)}{\|f_r(x)\|})$ forms a multi-level feature extractor and $W_C^b=concat(W^b_m,W^b_h,W^b_t)$ is the classifier weight matrix for base categories. Given a test image $x^b$, this model can be used to predict the label in the base label space $Y^b$, that is $argmax(M(f_C(x),W_C^b)) \in Y^b$.

\noindent\textbf{Generating weights for few-shot learning.} Now, we can utilize the feature extractor $f_C(\cdot)$ and weight matrix $W_C^b$ to construct different models for different few-shot learning settings. We first construct the weight matrix $W_C^n$ for $Y^n$ using a weight generator (AvgGen \cite{qi2018low} or AttGen \cite{gidaris2018dynamic}). Then, we can build classification models $M(f_C(x),W_C^n)$ and $M(f_C(x),[W_C^b,W_C^n])$ for standard and generalized few-shot learning scenario respectively. Here, the weight matrix $W_C^n$ is obtained by stacking each weight vector in order according to its label index in $Y^n$.

Let $Y^b$ and $Y^n$ denote the base- and novel-label space respectively,  we obtain its corresponding weight vector $w^y$ by normalizing the prototype of the given $k$ training samples $\{x^y_1,...,x^y_k\}$.

\begin{equation}
    w^y = \frac{\frac{1}{k}\sum_{i=1}^nf(x^y_i)}{\|\frac{1}{k}\sum_{i=1}^nf(x^y_i)\|},
\end{equation}
where $f(.)$ is the multi-level feature extractor derived from the combined base model.  Now, Given a unseen label space, we can build classification models $M(f(x),W^n)$ and $M(f(x),[W^b,W^n])$ for standard and generalized few-shot learning scenario respectively. Here, the weight matrix $W^n$ is obtained by stacking each weight vector in order according to its label index in $Y^n$, $W^b$ is weight matrix derived from the combined base model.

\section{Experiments}

\subsection{Datasets and evaluation metrics}

We validate our proposed method on Low-shot-ImageNet \cite{hariharan2017low} and Low-shot-CUB \cite{qi2018low} based on three performance metrics.

\noindent\textbf{Low-shot-ImageNet} contains 193 base categories,300 novel categories, 196 base categories, and 311 novel categories respectively. The first two groups are made for validating hyper-parameters, the remaining  two groups are used for the final evaluation.

\noindent\textbf{Low-shot-CUB} is constructed from Caltech-UCSD bird dataset \cite{wah2011caltech}. The dataset consists of 100 base classes and 100 novel classes. Since each category of this dataset contains only about 30 images, we repeat 20 experiments and take the average top-1 accuracy. 

\noindent\textbf{Performance evaluation metrics}. Few-shot learning methods are evaluated differently according to different few-shot learning setting.  These performance measures mainly differs in the way of constructing a test dataset. To evaluate our proposed method in both standard and generalized setting, we use three evaluation metrics summarized as below:

\noindent
\textbf{1) Novel/Novel:} the model's performance is measured by the accuracy of novel test examples within the novel label space, that is $D_{test} = \{(x_i,y_i)\in D_{test}^n,y_i \in Y^n \}$. \\
\textbf{2) Novel/All:} the model's performance is measured only by the accuracy of novel test examples in all label space, that is $D_{test} = \{(x_i,y_i)\in D_{test}^n,y_i \in Y^b \cup Y^n  \}$. \\
\textbf{3) All:} the model's performance is measured only by the accuracy of all test examples in all label space, that is $D_{test} = \{(x_i,y_i)\in D_{test}^b \cup D_{test}^n,y_i \in Y^b \cup Y^n  \}$.

Here,  standard few-shot learning setting only consider \textit{Novel/Novel} as the major performance measure, while the generalized setting consider results of both \textit{Novel/All} and \textit{All} . We report results of these metrics based on multiple tries. Specifically, in our experiments, we randomly select training images of the novel categories and repeat experiments 100 times, and finally report the mean accuracies within $95\%$ confidence intervals.





\subsection{Network architecture and training details}

\textbf{Network architecture}. We conduct experiments on the Few-shot-Imagenet benchmark using ResNet-10 and -50 \cite{he2016deep} architecture in our learning framework. For experiments on the Few-shot-CUB dataset, as Qi et al. \cite{qi2018low} obtained their results based on InceptionV1 \cite{szegedy2015going}, we implement our method based on the same network structure for performance comparison.

\noindent\textbf{Training details}. For all experiments on imageNet based few-shot benchmarks, we trained our model from scratch for 90 epochs on the base classes. The learning rate starts from 0.1 and is divided by 10 every 30 epochs with a fixed weight decay of 0.0001. We then fine-tune the model for further with the classifier-centric constraint with a small learning rate 0.0001.  For the CUB dataset experiment, all the pre-trained models we used are from the Pytorch official model zoo. During the training, the initial learning if 0.001 decreases by 0.1 times at 30 epoch intervals.

\subsection{Results and analysis}

\begin{table*}[ht!]
\begin{center}
\resizebox{\textwidth}{!}{
\begin{tabular}{l *{5}{c} |*{5}{c}| *{5}{c}}
\Xhline{2\arrayrulewidth}
 & \multicolumn{5}{c}{Novel / Novel} & \multicolumn{5}{c}{Novel / All}  & \multicolumn{5}{c}{All} \\
 Method & n=1 & 2 & 5 & 10 & 20 & n=1 & 2 & 5 & 10 & 20 & n=1 & 2 & 5 & 10 & 20 \\
 \hline
 
  
 Pro. Nets \cite{snell2017prototypical} (from \cite{agrawal2014analyzing}) & 39.4 & 54.4 & 66.3 & 71.2 & 73.9 & - & -& - & - & - & 49.5 & 61.0 & 69.7 & 72.9 & 74.6 \\
 
 Log. Reg. (from \cite{wang2018low}) & 38.4 & 51.1 & 64.8 & 71.6 & 76.6 & - & - & - & - & - & 40.8 & 49.9 & 64.2 & 71.9 & 76.9 \\
 
 Log. Reg w/G. (from \cite{wang2018low}) & 40.7 & 50.8 & 62.0 & 69.3 & 76.5 & - & - & - & - & - & 52.2 & 59.4 & 67.6  & 72.8 & 76.9 \\
 
 Pro. Mat. Nets \cite{wang2018low} & 43.3 & 55.7 & 68.4 & 74.0 & 77.0 & - & - & - & - & - & 55.8 & 63.1 & 71.1 & 75.0 & 77.1 \\
 
 Pro. Mat. Nets w/G \cite{wang2018low} & 45.8 & 57.8 & 69.0 & 74.3 & 77.4 & - & - & - & - & - & 57.6 & 64.7 & 71.9 & 75.2 & 77.5 \\
 
 SGM w/G. \cite{hariharan2017low} & - & - & - & - & - & 32.8 & 46.4 & 61.7 & 69.7 & 73.8 & 54.3 & 62.1 & 71.3 & 75.8 & 78.1 \\
 
 Batch SGM \cite{hariharan2017low} & - & - & - & - & - & 23.0 & 42.4 & 61.9 & 69.9 & 74.5 & 49.3 & 60.5 & 71.4 & 75.8 & 78.5 \\
 
 Mat. Nets \cite{vinyals2016matching}(from \cite{hariharan2017low,wang2018low}) & 43.6 & 54.0 & 66.0 & 72.5 & 76.9 & 41.3 & 51.3 & 62.1 & 67.8 & 71.8 & 54.4 & 61.0 & 69.0 & 73.7 & 76.5 \\

\multirow{2}{4.5cm}{Wei. Imprint* + AvgGen \cite{qi2018low}} & 44.05 & 55.42 & 68.06 & 73.96 & 77.21 & 38.70 & 51.36 & 65.89 & 72.60 & 76.21 & 56.73 & 63.66 & 71.04 & 74.05 & 75.47 \\
 & $\pm.21$ & $\pm.16$ & $\pm.09$ & $\pm.07$ & $\pm.05$ & $\pm.21$ & $\pm.17$ & $\pm.09$ & $\pm.07$ & $\pm.05$ & $\pm.13$ & $\pm.10$ & $\pm.06$ & $\pm.04$ & $\pm.03$ \\
 
\multirow{2}{4.5cm}{AvgGen (with retraining) \cite{gidaris2018dynamic}} & 45.23 & 56.90 & 68.68 & 74.36 & 77.69 & 39.33 & 50.27 & 63.16 & 69.56 & 73.47 & 54.65 & 64.69 & 72.35 & 76.18 & 78.46 \\
 & $\pm.25$ & $\pm.16$ & $\pm.09$ & $\pm.06$ & $\pm.06$ & $\pm.25$ & $\pm.16$ & $\pm.11$ & $\pm.07$ & $\pm.06$ & $\pm.15$ & $\pm.10$ & $\pm.06$ & $\pm.04$ & $\pm.04$ \\

\multirow{2}{4.5cm}{AttGen \cite{gidaris2018dynamic}} & 46.02 & 57.51 & 69.16 & 74.84 & 78.81 & 40.79 & 51.51 & 63.77 & 70.07 & 74.02 & 58.16 & 65.21 & 72.72 & 76.65 & 78.74\\
& $\pm.25$ & $\pm.15$ & $\pm.09$ & $\pm.06$ & $\pm.05$ & $\pm.25$ & $\pm.15$ & $\pm.12$ & $\pm.07$ & $\pm.06$ & $\pm.15$ & $\pm.09$ & $\pm.06$ & $\pm.04$ & $\pm.03$ \\
TRAML \cite{li2020boosting}~+~AttGen & 48.1 & 59.2 & 70.3 & 76.4 & 79.4 & - & - & - & - & - & 59.2 & 66.2 & 73.6 & 77.3 & 80.2 \\

\hline
\multirow{2}{4.5cm}{\textbf{MLWC~+~AvgGen}} & 48.22 & 58.77 & 69.71 & 74.45 & 76.91 & 44.06 & 55.83 & 68.15 & 73.36 & 76.07 & 58.96 & 65.18 & 71.28 & 73.63 & 74.78 \\
& $\pm.12$ & $\pm.09$ & $\pm.05$ & $\pm.03$ & $\pm.02$ & $\pm.12$ & $\pm.09$ & $\pm.05$ & $\pm.04$ & $\pm.02$ & $\pm.07$ & $\pm.05$ & $\pm.03$ & $\pm.02$ & $\pm.02$ \\
\multirow{2}{4.5cm}{\textbf{MLWC*~+~AvgGen}} & 49.09 & 59.66 & 70.26 & 74.72 & 77.04 &45.56& 57.12 & 68.85 & 73.73 & 76.24 & 59.37 & 65.48 & 71.36 & 73.63 & 74.72 \\
& $\pm.11$ & $\pm.08$ & $\pm.04$ & $\pm.03$ & $\pm.02$ & $\pm.11$ & $\pm.09$ & $\pm.05$ & $\pm.03$ & $\pm.02$ & $\pm.07$ & $\pm.05$ & $\pm.03$ & $\pm.02$ & $\pm.02$ \\
\multirow{2}{4.5cm}{\textbf{MLWC*~+~AttGen}} & \textbf{50.87} & \textbf{62.13} & \textbf{72.61} & \textbf{77.02} & \textbf{79.67} & \textbf{46.18} & \textbf{57.21} & \textbf{68.63} & \textbf{73.64} & \textbf{76.59} &\textbf{61.72} & \textbf{68.58} & \textbf{75.35} & \textbf{78.29} & \textbf{80.03} \\
& $\pm.22$ & $\pm.15$ & $\pm.09$ & $\pm.06$ & $\pm.23$ & $\pm.15$ & $\pm.09$ & $\pm.09$ & $\pm.07$ & $\pm.05$ & $\pm.14$ & $\pm.08$ & $\pm.06$ & $\pm.05$ & $\pm.03$ \\

 \Xhline{2\arrayrulewidth}
\end{tabular}
}
\end{center}
\caption{Comparison with the state-of-art methods using Resnet-10 on the Low-shot-ImageNet dataset. Best are bolded. * indicates that we get 5 random crops from each training example, then use the average feature as the weight of novel class.}
\label{table:imagenet}
\end{table*}

\begin{table*}[ht!]
\begin{center}
\resizebox{\textwidth}{!}{
\begin{tabular}{l *{5}{c} |*{5}{c}| *{5}{c}}
\Xhline{2\arrayrulewidth}
 & \multicolumn{5}{c}{Novel / Novel} & \multicolumn{5}{c}{Novel / All}  & \multicolumn{5}{c}{All} \\
 Method & n=1 & 2 & 5 & 10 & 20 & n=1 & 2 & 5 & 10 & 20 & n=1 & 2 & 5 & 10 & 20 \\
 \hline
 Mat. Nets \cite{vinyals2016matching} (from \cite{wang2018low}) & 53.5 & 63.5 & 72.7 & 77.4 & 81.2 & - & - & - & - & - & 64.9 & 71.0 & 77.0 & 80.2 & 82.7 \\
 
 Pro. Nets \cite{snell2017prototypical}  & 49.6 & 64.0 & 74.4 & 78.1 & 80.0 & - & - & - & - & - & 61.4 & 71.4 & 78.0 & 80.0 & 81.1 \\
 
 Pro. Mat. Nets w/G \cite{wang2018low} & 54.7 & 66.8 & 77.4 & 81.4 & 83.8 & - & - & - & - & - & 65.7 & 73.5 & 80.2 & 82.8 & 84.5 \\
 SGM w/G. (from \cite{wang2018low})& - & - & - & - & - & 45.1 & 58.8 &72.7 & 79.1 &82.6 & 63.6 & 71.5 & 80.0 & \textbf{83.3} & \textbf{85.2} \\
 \hline
 \multirow{2}{4.5cm}{\textbf{MLWC + AvgGen}} & 57.12 & 68.28 & 77.77 & 81.80 & 83.72 & 53.48 & 65.05 & 76.59 & 80.95 & 83.07 & 67.49 & 73.36 & 79.87 & 81.98 & 82.95 \\
& $\pm.20$ & $\pm.14$ & $\pm.07$ & $\pm.07$ & $\pm.04$ & $\pm.23$ & $\pm.13$ & $\pm.08$ & $\pm.08$ & $\pm.04$ & $\pm.14$ & $\pm.08$ & $\pm.05$ & $\pm.05$ & $\pm.02$ \\

\multirow{2}{4.5cm}{\textbf{MLWC*+AvgGen}} & \textbf{57.97} & \textbf{69.08} & \textbf{78.19} & \textbf{81.99} & \textbf{83.80} & \textbf{54.82} & \textbf{66.93} & \textbf{77.12} & \textbf{81.22} & \textbf{83.16} & \textbf{68.01} & \textbf{74.72} & \textbf{79.98} & 81.99 & 82.88 \\
& $\pm.20$ & $\pm.15$ & $\pm.06$ & $\pm.07$ & $\pm.03$ & $\pm.22$ & $\pm.05$ & $\pm.05$ & $\pm.08$ & $\pm.03$ & $\pm.13$ & $\pm.09$ & $\pm.05$ & $\pm.05$ & $\pm.02$ \\

 \Xhline{2\arrayrulewidth}
\end{tabular}
}
\end{center}
\caption{Comparison with the state-of-art methods using Resnet-50 on the Low-shot-ImageNet dataset.Best are bolded. * indicates that we get 5 random crops from each training example, then use the average feature as the weight of novel class.}
\label{table:imagenet2}
\end{table*}

\subsubsection{Low-shot Classification accuracy} 
We evaluated the performance of the proposed method on two low-shot benchmarks. 

\noindent\textbf{Low-shot-ImageNet.} Table \ref{table:imagenet} and \ref{table:imagenet2}  provide the comparative results of different techniques using two network backbones on the large-scale Few-shot-ImageNet dataset. First, we can observe that some existing methods show significant improvement on one evaluation metric but minor on another one. For example, both Weight imprinting \cite{qi2018low} and AttGen \cite{gidaris2018dynamic} have better performance than Matching Nets \cite{vinyals2016matching} in the "Novel/Novel" setting but similar or even worse results in the "Novel/ALL" setting. In comparison, our approach consistently achieves the best results on all evaluation metrics.  Specifically,  using the same weight generator AttGen,  our method significantly outperforms the current best model TRAML \cite{li2020boosting} in testing both novel-class and all-class classification accuracy. Besides, without learning the weight generator, our proposed method also achieves a comparable performance to the current top-performing methods that require training a weight generator. For instance, compared to the TRAML method that needs to learn an attention-based weight generator, our approach obtains a similar performance using the mean feature as classifier weights. All these results indicate that our learned representation yields a better generalization ability and versatility for FSL learning.  

\begin{table*}[ht!]
\begin{center}
\resizebox{\textwidth}{!}{
\begin{tabular}{l *{5}{c} |*{5}{c}| *{5}{c}}
\Xhline{2\arrayrulewidth}
 & \multicolumn{5}{c}{Novel / Novel} & \multicolumn{5}{c}{Novel / All}  & \multicolumn{5}{c}{All} \\
  
 Method & n=1 & 2 & 5 & 10 & 20 & n=1 & 2 & 5 & 10 & 20 & n=1 & 2 & 5 & 10 & 20 \\
 
 \hline
Gen. + Cla.\cite{hariharan2017low} )from \cite{qi2018low}) & - & - & - & - & - & 18.56 & 19.07 & 20.00 & 20.27 & 20.88 & 45.42 & 46.56 & 47.79 & 47.88 & 48.22 \\

Mat. Nets \cite{vinyals2016matching}(from \cite{qi2018low})  & - & - & - & - & - & 13.45 & 14.75 & 16.65 & 18.18 & 25.77 & 41.71 & 43.15 & 44.46 & 45.65 & 48.63 \\
Imprinting  \cite{qi2018low} & - & - & - & - & - & 21.26 & 28.69 & 39.52 & 45.77 & 49.32 & 44.75 & 48.21 & 52.95 & 55.99 & 57.47 \\
Imprinting* \cite{qi2018low} & - & - & - & - & - & 21.40 & 30.03 & 39.35 & 46.35 & 49.80 & 44.60 & 48.48 & 52.78 & 56.51 & 57.84 \\
\hline
\textbf{MLWC}        &32.35 & 39.78 & 49.47 & 54.67 & 57.37 &30.72 & 37.65   & 48.17 & 53.56 & 56.45  & 49.80  & 53.41 & 57.87 & \textbf{60.46} & \textbf{61.61}  \\
\textbf{MLWC*}  &\textbf{33.56} & \textbf{40.82} & \textbf{50.28} & \textbf{54.67} & \textbf{57.53} & \textbf{30.87} & \textbf{39.01}  & \textbf{49.17} & \textbf{53.66} & \textbf{56.61}  & \textbf{49.96}  & \textbf{53.73} & \textbf{58.18} & 60.30 & 61.60   \\
\hline
\end{tabular}
}
\end{center}

\caption{Comparison with the state-of-art methods on the Few-shot-Cub dataset. * indicates that we get 5 random crops from each training example, then use the average feature as the weight of novel class.}
\label{table:cub}
\end{table*}

\begin{table*}[ht!]
\begin{center}
\resizebox{\textwidth}{!}{
\begin{tabular}{l *{5}{c} |*{5}{c}| *{5}{c}}
\Xhline{2\arrayrulewidth}
 & \multicolumn{5}{c}{Novel / Novel} & \multicolumn{5}{c}{Novel / All}  & \multicolumn{5}{c}{All} \\
  
 Method & n=1 & 2 & 5 & 10 & 20 & n=1 & 2 & 5 & 10 & 20 & n=1 & 2 & 5 & 10 & 20 \\
 
 \hline
 Imprinting* \cite{qi2018low}(Resnet50*) &32.15  & 40.48 & 52.41 & 57.93 & 61.72 & 26.24 & 35.79 & 49.31 & 55.31 & 59.38 & 52.43 & 56.83 &  62.89 & 65.53& 67.27 \\
\textbf{MLWC} & 35.91 & 44.91 & 56.95 & 62.48 & 66.01 & 33.54 & 43.47 & 56.21 & 61.96 & 65.61 & 55.45 &  59.58 & 64.94 & 67.32& 68.78 \\
\textbf{MLWC*} & \textbf{36.96} & \textbf{45.53} & \textbf{57.43} & \textbf{63.03} & \textbf{66.35} & \textbf{34.91} & \textbf{44.21} & \textbf{56.81} & \textbf{62.52} & \textbf{65.96} & \textbf{55.60} &  \textbf{59.66} & \textbf{65.02} & \textbf{67.46}& \textbf{68.89} \\
\hline
\end{tabular}
}
\end{center}

\caption{Comparison with the state-of-art methods on the Few-shot-Cub dataset. * indicates that we get 5 random crops from each training example, then use the average feature as the weight of novel class.}
\label{table:cub1}
\end{table*}

\noindent\textbf{Low-shot-CUB.} Since existing method reported on this dataset is based on Inception V1 network, we first evaluate our method with the same backbone network.  Table \ref{table:cub} shows performance comparison result of different approaches.  Our proposed method outperforms all the comparing method by a large margin in all evaluation metrics. For instance, our method achieves top-1 accuracies of 30.72\% and 37.65\% under the 1 and 2 shot settings respectively, the previous best results are 21.40\% and 28.69\%. To evaluate out method's effectiveness on this dataset when using different network architecture, we  further use the Resnet-50 as backbone for both  the Imprinting and our method and compare their performance.  Table \ref{table:cub1} shows the corresponding results and, again, demonstrates our method's superior performance in low-shot learning.

 \begin{table*}[!]
\begin{center}
\resizebox{\textwidth}{!}{
\begin{tabular}{l *{5}{c} |*{5}{c}}
\Xhline{2\arrayrulewidth}
 & \multicolumn{5}{c}{Novel classes from ImageNet} & \multicolumn{5}{c}{Novel classes from CUB2011}   \\
  
 Method & n=1 & 2 & 5 & 10 & 20 & n=1 & 2 & 5 & 10 & 20 \\
 
 \hline
High-level(baseline) & 51.56 & 63.67 & 74.78 & 79.68 & 82.45 & 30.55 & 40.76 & 53.68 & 60.79 & 65.54 \\
Mid-level & 51.59 & 63.80 & 75.57 & 80.60 & 83.21 & 35.99 & \textbf{48.40} & \textbf{62.51} & \textbf{70.26} & \textbf{74.92} \\
Relation-level & 48.94 & 58.64 & 69.23 & 73.32 & 75.65 & 24.45 & 32.19 & 40.92 & 46.18 & 49.18 \\
\textbf{Multi-level}  & \textbf{55.50} & \textbf{67.51} & \textbf{78.26} & \textbf{82.75} & \textbf{85.00} & \textbf{36.15} & 48.34 & 62.44 & 69.94 & 74.37 \\

\hline

\end{tabular}
}
\end{center}
\caption{The performance of using different levels of representation for few-shot learning on the same task (Generic object classification) and another different task (Fine-grained object classification). Top-5 accuracy of the novel categories in the novel label space (Novel/Novel) is reported. Best are bolded.}
\label{table:crosstask}
\end{table*}

\noindent\textbf{Cross-domain performance of low-shot learning}. We investigate the transferability of different levels of representations in the FSL setting. To achieve this, we perform a cross-domain evaluation, where we evaluate the learned model on both the same-domain and different domain data. Specifically, we first train a model on the base-class data from the ImageNet dataset. Then we evaluate it on both the ImageNet and the  Caltech-UCSD bird dataset \cite{wah2011caltech}. Table \ref{table:crosstask} presents the comparison results obtained based on the Resnet-50 backbone and the Avg weight generator. First, we can observe that the mid-level features achieve the best accuracy in cross-domain testing while the relation-level performs the worst. This result reveals that the mid-level representation exhibit strong transferability in the FSL setting. Furthermore, the proposed multi-level representation achieves the best accuracy on the same-domain data and obtains comparable performance with the mid-level features. This indicates that using multi-level features for FSL help improve generalization ability and handle domain shift problem.

\subsubsection{Analysis and ablation study}

\begin{table}
\begin{center}

\resizebox{\textwidth}{!}{
\begin{tabular}{l *{5}{c}|c}
\Xhline{2\arrayrulewidth}
 Method & n=1 & 2 & 5 & 10 & 20  &  Classifier\\
\hline
Baseline  \cite{qi2018low,gidaris2018dynamic} & 53.96 & 62.88 & 69.55 & 71.56 & 73.42 & 81.80  \\
\textbf{Baseline + WC} & \textbf{69.93} & \textbf{74.94}  & \textbf{78.30} & \textbf{78.99} & \textbf{79.68} & 81.71 \\
\hline
\end{tabular}
}
\end{center}
\caption{Classification accuracy of CUB validation set using samples as the classifier in two feature spaces. }
\label{table:bacc}
\end{table}

\begin{figure*}[t]
\begin{center}
  \includegraphics[width=1\linewidth]{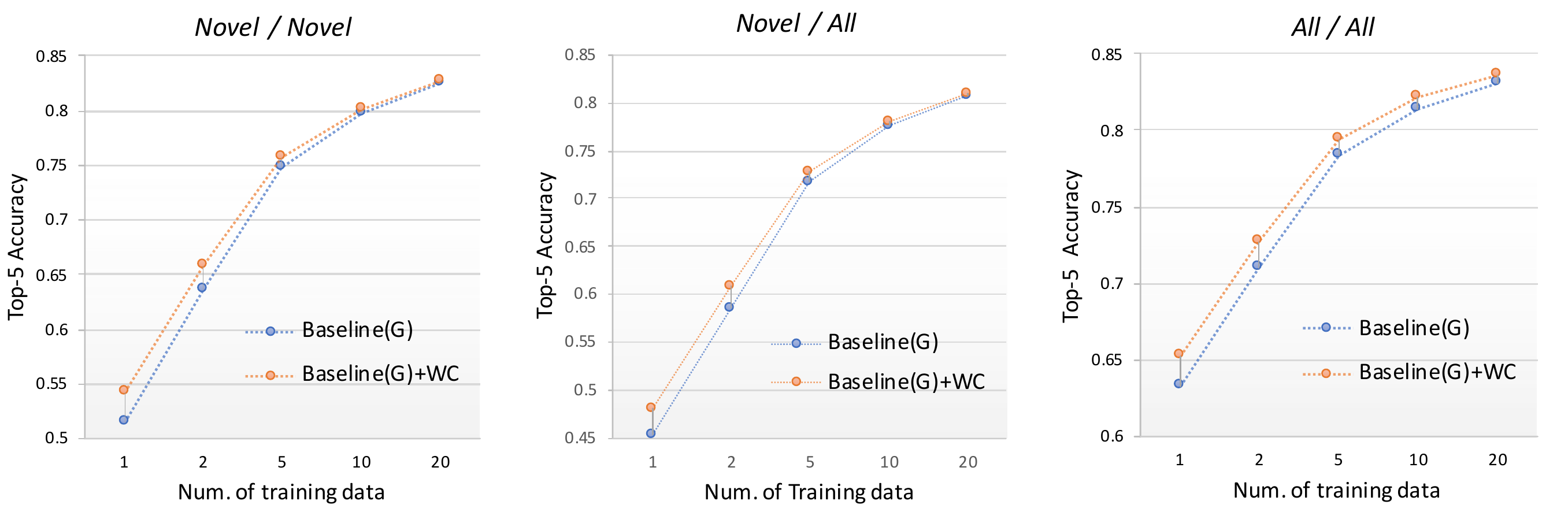}
\end{center}
   \caption{Comparison of the inter-class variance between two feature spaces both learned on base training set. Here, baseline refers to the feature space learned with cosine softmax loss, WC denotes our proposed weight-centric constrain. Noted that we report the average inter-class variance for each dataset. }
\label{fig:fewshotacc}
\end{figure*}

\noindent\textbf{Effectiveness of the classifier-centric constraint}. To verify the effectiveness of the classifier-centric constraint, we established the following experiments. First, we train two ConvNet models on the base class data, with and without classifier-centric constraints to learn the two feature spaces. Then we randomly sample some samples from each class of the base class dataset to construct two classifiers to classify the test set. Finally, by evaluating their classification performance, it is indicated in which feature space the sample can construct a better decision boundary. The experimental results are shown in Tab.\ref{table:bacc}. We can observe that the feature space learned with cosine softmax loss achieve poor accuracy, that indicates the sample points in this space might be scattered and not close to the classifier weight. By applying the  classifier-centric constraint, the accuracy is significantly improved. This demonstrates that the feature space learned with classifier-centric constraint is more suitable for building classifiers using samples. We further evaluate the classifier-centric constraint under different evaluation metrics and provide the results in Fig.\ref{fig:fewshotacc}. We can see that our proposed constraint improves the baseline consistently in three evaluation metrics. More importantly, the improvements under the "ALL/ALL" setting is the most significant, which reveals that the classifier-centric constraint exhibits superiority in the generalized few-shot learning. 
   \begin{table}[!]
\begin{center}

\resizebox{\textwidth}{!}{
\begin{tabular}{l *{3}{c} |*{3}{c}}
\Xhline{2\arrayrulewidth}
 & \multicolumn{3}{c}{Novel / Novel} & \multicolumn{3}{c}{Novel / All} \\
  & n=1 & 2 & 5 & n=1 & 2 & 5  \\
 \hline
H(baseline) & 51.56 & 63.67 & 74.78 & 45.26 & 58.53 & 71.80 \\
\hline
H+WC & 54.24 & 65.71 & 75.75 & 47.95 & 60.77 & 72.91 \\
(H+WC)+M & 56.96 & 68.50 & 78.58 & 50.79& 64.30 & 76.52 \\
(H+WC+M)+R & 57.12 & 68.28 & 77.77 & 53.48 & 65.82 & 76.95 \\
\hline
\end{tabular}}

\end{center}
\caption{Oblation study experiments on the ImageNet based few-shot benchmark. \textit{H}, \textit{M},and \textit{R} refer to High-, Mid-, and relation-level features, respectively. \textit{WC} refers to using weight-centric learning strategy.}
\label{table:oblation}
\end{table}
\begin{figure*}
\begin{center}
\includegraphics[width=0.85\linewidth]{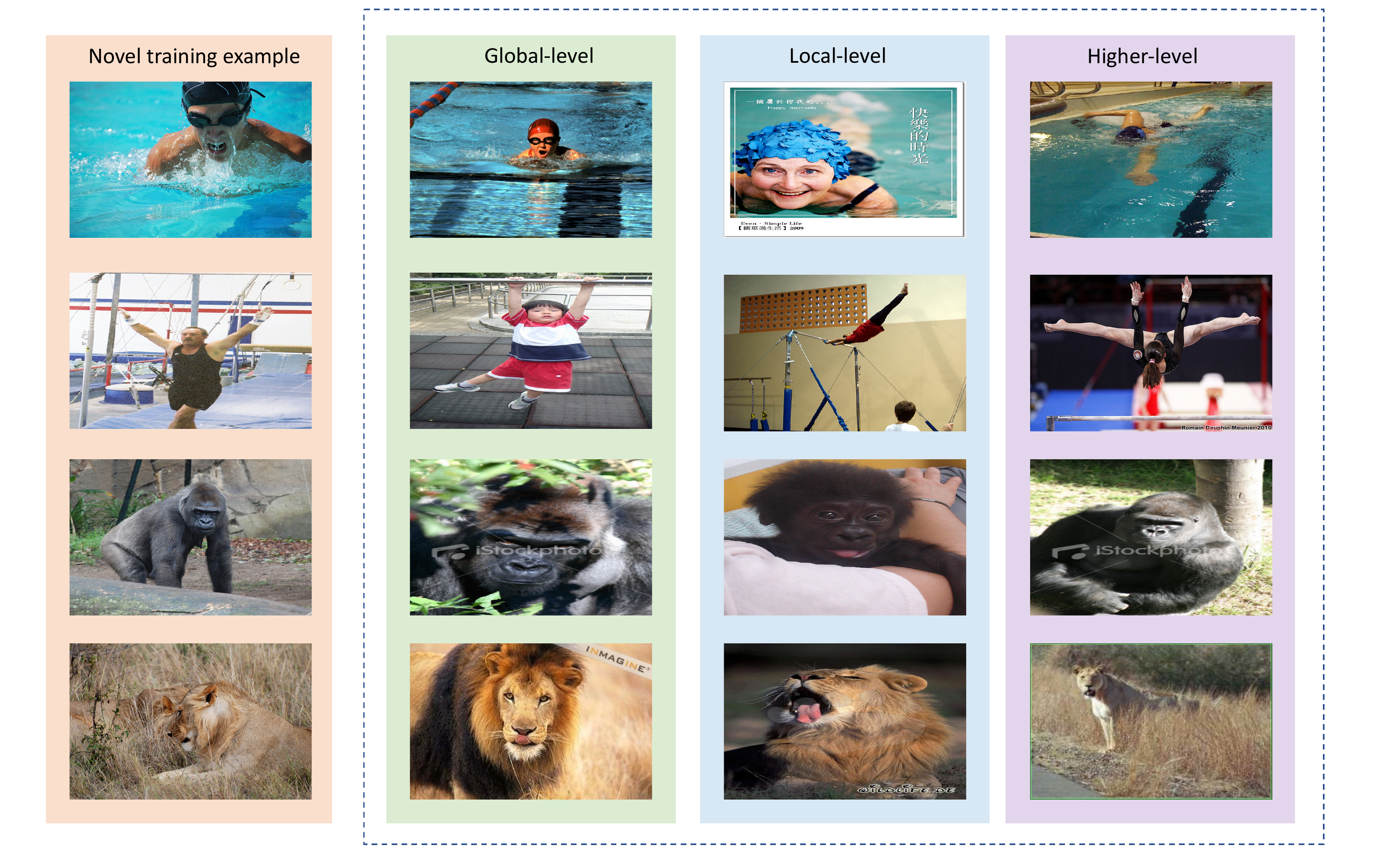}
\end{center}
\caption{Some successful exemplars using our proposed method. The first column shows a single training image of novel class, all images in the remaining three columns are correctly predicted by using  the proposed multi-level representation. The second column shows some successful predictions using only global-level features but they are misclassified if using local or higher-level representation, and so on for the second and the third column. }
\label{fig:prediction}

\end{figure*}

\noindent\textbf{The contribution of each component.} We conduct ablation study to compare the performance of different levels representation in the FSL setting. Tab. \ref{table:oblation} provides an ablation study on the Few-shot-imagenet benchmarks to observe the effect of each element.  On the one hand, we can see that when evaluating only the novel label space,  adding the weight-centric and mid-level component in sequence continuously improves the performance. This demonstrates that both pieces help enhance the model generalization ability, which also implies that increasing the prototype-ability and transferability of feature representation can benefit few-shot learning. On the other hand, incorporating relation-level features does not further raise the performance in this setting. However,  it shows a significant improvement under the "novel/all" evaluation metric. This indicates that the relation-level features have weaker generalization to novel classes but can effectively prevent novel-class data from being classified into the base categories.




We also provide some prediction results in Fig. \ref{fig:prediction}, which can be used to intuitively analyze the few-shot learning ability of different representation. For example, the test images in the second column mostly contain some patterns (e.g., objects or parts of objects) which are very similar to those occurs in the training examples, while the similarities between images in the last two columns and the training images tend to be subtle.


\section{Conclusion}
This work aims to learn a good feature representation for FSL. We first complement the existing method's representation capacity by incorporating mid- and relation-level features. Then we introduce a classifier-centric learning strategy that allows a few sample features to construct a more discriminative classifier. We extensively evaluate our approach on two low-shot classification benchmarks and demonstrate its effectiveness in improving generalization. Our proposed method can also benefit other tasks such as zero-shot learning and image retrieval, in which feature extractors play a critical role. However, this work constructed multi-level feature by concatenating multiple features, which might bring much redundancy in features.  Therefore, we will investigate how to learn a compact representation from multiple information sources in future work. 

\bibliography{main.bib}

\end{document}